\documentclass[journal]{IEEEtran}
\usepackage{graphicx}
\usepackage{bm}
\usepackage{algorithm}
\usepackage{algorithmic}
\usepackage{multirow}
\usepackage{amsmath}
\usepackage{xcolor}
\usepackage{graphics}
\usepackage{graphicx}
\usepackage{float}
\usepackage{mathtools}
\usepackage{amsmath,amsthm}
\usepackage{amssymb}
\usepackage{multirow}
\usepackage{epstopdf}
\usepackage{caption}
\usepackage{fancyhdr}
\usepackage{paralist}
\usepackage{setspace}
\usepackage{stfloats}
\usepackage{array,color}
\usepackage{enumerate}
\usepackage{threeparttable}
\usepackage{cite}
\usepackage{subcaption}

\ifCLASSINFOpdf

\else

\fi

\begin{document}

\title{An Adaptive Oversampling Learning Method for Class-Imbalanced Fault Diagnostics and Prognostics}

\author{\IEEEauthorblockN{Wenfang Lin, Zhenyu Wu, Yang Ji}
\IEEEauthorblockA{}}

\maketitle

\begin{abstract}
Data-driven fault diagnostics and prognostics suffers from class-imbalance problem in industrial systems and it raises challenges to common machine learning algorithms as it becomes difficult to learn the features of the minority class samples. Synthetic oversampling methods are commonly used to tackle these problems by generating the minority class samples to balance the distributions between majority and minority classes. However, many of oversampling methods are inappropriate that they cannot generate effective and useful minority class samples according to different distributions of data, which further complicate the process of learning samples. Thus, this paper proposes a novel adaptive oversampling technique: EM-based Weighted Minority Oversampling TEchnique (EWMOTE) for industrial fault diagnostics and prognostics. The methods comprises a weighted minority sampling strategy to identify hard-to-learn informative minority fault samples and Expectation Maximization (EM) based imputation algorithm to generate fault samples. To validate the performance of the proposed methods, experiments are conducted in two real datasets. The results show that the method could achieve better performance on not only binary class, but multi-class imbalance learning task in different imbalance ratios than other oversampling-based baseline models.
\end{abstract}


%
\IEEEpeerreviewmaketitle

\section{Introduction}

Fault diagnostics and prognostics becomes important and pervasive in industrial field since the large volume of industrial data being collected from the industrial process, and industrial systems' performance could be monitored and predicted. It helps operators prevent unexpected accidents, decrease downtime and save costs [1]. For example, in electricity generation system, wind turbines blade usually suffers from freezing problem [2]. Accurately predicting when freezing occurs is vital for wind turbine maintenance, which can decrease the fault risks and save costs. At the same time, in terms of the factory assembly line, the entire production process will be blocked owing to any error in one link, early detection of faults can avoid system shut-down and component failure or even catastrophes [3].

At present, fault diagnostics and prognostics for specific industrial scenarios (e.g., aircraft systems, wind turbines, motors) are usually based on a mathematical model of the system. however, the performance of a model-based diagnostics and prognostics scheme depends on the accuracy of the mathematical model, which is difficult to obtain and is influenced by the assumptions and rules [4]. At the same time, it is difficult to establish a reasonable mathematical model by the mechanism without expert knowledge about this filed, especially a complex industrial system. Therefore, in recent years, data-driven techniques have been widely used in fault diagnostics and prognostics for specific industries. This paper also focuses on data-driven techniques in the filed of different industrial systems.

Data-driven fault diagnostics and prognostics systems have higher requirements on samples, and a sufficient number and well-defined samples are more helpful for model learning. At the same time, the number of samples for each class(e.g. normal samples and some fault samples) also plays a crucial role. However, real-life industrial systems usually operate under normal conditions, and few faults occur, which may result in the collected data being mostly unbalanced data, and most fault diagnostics and prognostics classifiers are designed based on balanced samples, which further makes the process of learning the samples and detecting faults complicated. Although there are many data-driven techniques to diagnose industrial systems faults, there is still a need to design data-driven techniques to diagnose industrial systems problems under the class imbalanced conditions. Therefore, in the paper, we design a general data-driven technique for fault diagnostics and prognostics for class imbalanced learning.

Several popular methods can handle class imbalanced learning problems, one is the use of sampling techniques, which can balance distributions by adding the minority samples or removing the majority samples, and the other method to tackle the problems is improvement on classification algorithms. This paper focuses on the data-level method to tackle the class imbalanced learning problems that can be used by the most of the fault classifiers. Therefore, we studied the advantages and disadvantages of simple sampling methods(e.g, random downsampling and random oversampling), as well as advanced sampling techniques about the synthetic oversampling methods(e.g, SMOTE [5]).

This paper proposes an general scheme of imbalanced fault disgnostics and prognostics in industrail systems, which consist of the state-of-the-art techniques about feature extraction, feature reduction and data-level oversampling. We are committed to selecting the most appropriate methods about feature extraction, feature reduction and data-level oversampling method in different industrial scenes. Meawhile, this paper proposes an efficient adaptive synthetic oversampling technique EWMOTE (EM-based Weighted Minority Oversampling Technique) for industrial fault diagnostics and prognostics, which generates effective minority class examples to alleviate the imbalanced learning problems. This proposed technique improves the MWMOTE(Majority Weighted Minority Oversampling Technique) [6] algorithm, which utilizes the rules of MWMOTE to identify the hard-to-learn informative minority class samples and missing data imputation by EM(Expectation Maximization Imputation)[7] method to generate effective samples with the distribution of overall minority samples.

In order to verify the validity of the proposed oversampling algorithm, we have compared the performance on two challenges' datasets, a wind turbine freezing failure forecast task and a plant's fault prediction task. The attained results show that the proposed algorithm outperforms other state-of-the-art oversampling methods in fault diagnostics and prognostics on both binary fault prediction and multi-faults classification task with multiple different imbalance ratio.
%

The remainder of the paper is organized as follows: several state-of-arts studies on fault diagnostics and prognostics in industrial fields and data-level class-imbalance learning methods are surveyed in section II. In Section III, we formally describe an overview of the imbalanced fault industrial diagnostic and prognostics
system, and the proposed EWMOTE algorithm is illustrated in detail. Section IV introduces experiments and evaluations on two challenges' datasets to validate the feasibility and effectiveness of proposed model. Finally, conclusions are drawn and future work is presented in section V.

\section{Related Work}

In general, fault diagnostics and prognostics methods can be categorized into data-driven, model-based and hybrid approaches. Data-driven approaches use training data to learn the characteristics of the normal state and different fault states. Model-based approaches build a mathematical model that describes the fault or normal state, and combine the mathematical model with data to identify model parameters and states of industrial systems. Hybrid approaches combine the above methods to improve the performance of fault diagnostics and prognostics.

Data-driven approaches generally are based on machine learning to tackle the fault diagnostics and prognostics problems, for example, Bayesian network [9] and artificial neural networks (ANNs) [10] are two useful methods that have been used in railway traction device and grinding mill liners, respectively. Gaussian process (GP) regression [11], relevance/support vector machine [12,13] and fuzzy-logic [14]are also widely used to tackle the fault diagnostics and prognostics problems. Meanwhile, some ensemble machine learning approaches (e.g., random forest and gradient boosted tree) [15] are used to fault detection in aircraft systems. These works in industrial fault diagnostics and prognostics mostly focus on how to perform feature selection and improve the machine learning algorithm based on the specific scenarios, and ignore the impact on distribution of training samples, especially on class imbalance samplings. So, some commonly-used class-imbalance learning techniques should be surveyed and applied to this field.

Several techniques have been proposed to tackle the class-imbalance learning problems in different fields. Class-imbalance learning techniques can be divided into three categories: sampling-based methods and algorithm-level methods, ensemble-based methods. In this paper, we mainly focus on the most efficient class imbalanced learning techniques for fault diagnostics and prognostics under class imbalance conditions.

Some sampling-based methods, including oversampling, undersampling and hybrid methods, are used for fault diagnostics and prognostics, which are more versatile because they are independent of the selected classifier [16]. Xie at al. [8] showed that oversampling is usually more useful than undersampling methods. Some synthetic techniques can generate new samples and add potential information to the original data, which can avoid the overfitting and improve classifiers' performance. For example, SMOTE [5] is a synthetic technique, which can add new minority class examples by k-nearest neighborhood. There are also some improved methods based on SMOTE, such as Borderline-SMOTE [17]. Some cluster-based oversampling methods have been proposed to partition the dataset and oversampling on different partitions' data [18,19]. Roozbeh et al. [7] proposed a novel imputation-based oversampling technique EMICIL for class imbalance learning. In order to generate more effective and useful minority class samples, some works chose proper base samples to generate new minority class samples. Adaptive synthetic (ADASYN) [14] and CBSO [15] try to tackle these problems in imbalanced learning problems. These methods based on the idea of assigning weights to minority class samples according to their importance. Sukarna Barua et al.[16] analysis the shortcomings of the approaches based on [20,21] cannot effectively assign weights to the minority class samples, and proposed ProWSyn [22] and MWMOTE [6] to generate effective weight values for the minority data samples based on samples proximity information. Adiition, some hybrid methods[23][24] ensembled the oversampling and undersampling can also improve the accuracy.


Algorithm-level methods aim at enhancing a classifier's performance based on its inherent characteristics in fault diagnostics and prognostics under class imbalance conditions. Modified kernel Fisher discriminant analysis [25], extreme learning machine [26] and some improved SVM(e.g., FSVM-CIL [27], GSVM [28]) have been widely used to tackle the imbalance problems in industrial fields. Meanwhile, some state-of-the-art ensemble-based class imbalanced learning approaches improve the performance by combining the Algorithm-level methods or sampling-based methods, including EasyEnsemble [8], RankCost [29], BNU-SVMs [30] and so on.


\section{Problem Formulation}

Two use cases and datasets are introduced in this section to help with formulating the problems.

\textbf{Use Case 1: Wind Turbine Freezing Failure Forecast}, the dataset$^1$\footnotetext[1]{Challenge datasets: https://github.com/minelabwot/ImbalanceLearning} is provided by a PHM competition$^2$\footnotetext[2]{Challenge homepage: http://www.industrial-bigdata.com} held by MIIT of China. The data contains 28-dimensional time series, including working conditions, environment parameters and state parameters of two wind turbines from their SCADA system. The data are labelled with normal and freezing durations with start\_time and end\_time, and the challenge is to predict the the duration of freezing failures from given test dataset.

\textbf{Use Case 2: Industrial Plant Failure Detection}, the dataset is provided by the PHM Society 2015 Data Challenge $^3$\footnotetext[3]{Challenge homepage: https://www.phmsociety.org/}, which is generated from SCADA of the industrial plant. The data contains 8-dimensional time series, including 4 sensor measurements and 4 control reference signals of industrial plants. The data are labelled with 6 faulty event types. The objective of the challenges is to detect the failure events of type 1-6 from given test data.

\begin{algorithm}
\caption{Fault Diagnostics and Prognostics Formulation}
\label{alg:2}
    \begin{algorithmic}[1]
        \REQUIRE $S = (E_1,E_2,E_3...,E_n)$, $E_i = (t^{start}, t^{end} ,label)$,   $i={1,2,...,n}$, $S'= $ $\emptyset$
        \STATE Train a classier $lable = F(E_i)$ with labeled $E_i$ in $S$
        \STATE Identify the $label$ of unlabeled $E_x$ in $S$ by classifier $F$
        \REPEAT
        \FOR {all $g\in$ $S$}
            \FOR {all $h\in$ $S$}
                \IF {$g$ and $h$ overlap and have the same label}
                    \STATE Combine $g$ and $h$ into a single event $E'$
                    \STATE Delete $g$ and $h$ from $S$, add $E'$ into $S$
                \ENDIF
            \ENDFOR
        \ENDFOR
        \UNTIL No overlapping events with same label are found
        \STATE $S'= S$
        \ENSURE The prediction of fault $S' = ({E_1}',{E_2}',...,{E_m}')$
    \end{algorithmic}
\end{algorithm}

According to the description of use cases, the problem could be categorized into fault diagnostics and prognostics respectively. Fault diagnostics refers to the process of fault detection and isolation, and fault prognostics refers to predict the time when the fault occurred and ended. Both of the two problem could be formulated as classification problems. The idea is that the original multi-dimensional signals $\boldsymbol{Sig}=(\boldsymbol{x_0},\boldsymbol{x_1},...,\boldsymbol{x_t})$ (supposed totally t timestamps) can be segmented into time windows $E_i (i={1,2,...,n})$ by sliding window, and each window $E_i$ could be labeled as normal or faulty event, which represent the machine's status at time $i$. The failure duration is modeled as continuous events with same faulty labels. In this way, the fault diagnostics is to identify the fault types of $E_t$ at given time $t$ from given test data. While, the $start\_time$ and $end\_time$ of failures $S' = ({E_1}',{E_2}',...,{E_m}')$, ${E_i}' = (t^{start}, t^{end} ,label)$, $i={1,2,...,m}$ , are able to be predicted [17]. The detailed process of problem formulation is introduced in Algorithm 1.

\section{Methodology}


In this section, the general scheme of tackling the imbalanced fault diagnostics and prognostics is proposed in Fig 1. The scheme consist of five main units: Raw Data, Segmentation, Feature Extraction and Feature Reduction, Data-level methods unit and Learning unit. The Data-level methods unit includes None-Sampling and other four oversampling methods. The learning unit includes fault detection and prediction. In this paper, we mainly focus on data-level methods, each unit will be introduced in details as in following subsections.

\begin{figure}[htbp]
\vspace{0pt}
\begin{center}
  \includegraphics[width=0.5\textwidth]{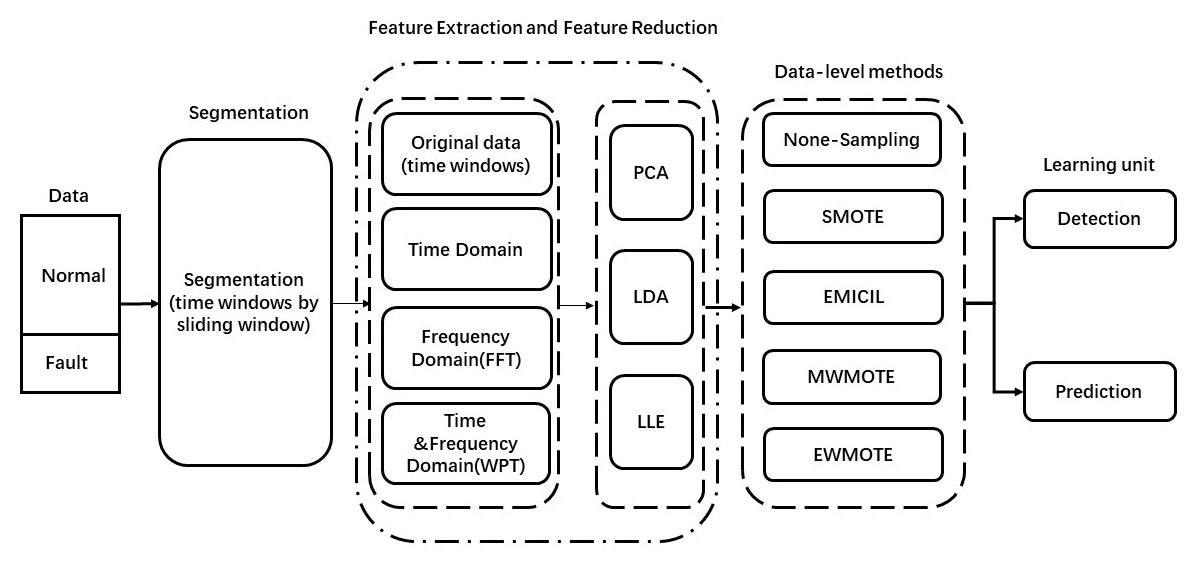}\\
  \caption{The general scheme of imbalanced fault diagnostics and prognostics in industrail Systems}\label{Figure 1}
\end{center}
\vspace{-10pt}
\end{figure}

\subsection{Segmentationm, Feature Extraction and Featurn Reduction}
Before putting the multi-dimensional time series into the training model, raw data should be segmented into time windows by sliding window. The length of time window is denoted as $L$ and sliding length as $N$. For n-dimensional time series at the time $t$, the feature could be denoted as $\boldsymbol{F_t}=\{\boldsymbol{x_1},\boldsymbol{x_2},\ldots,\boldsymbol{x_n}\}^T$, and the time window feature at $t$ is defined as $\boldsymbol{E_t}=\{\boldsymbol{F_t},\boldsymbol{F_{t+1}},\ldots,\boldsymbol{F_{t+L-1}}\}$, which is a stacked features to show the window feature. The description about feature extraction is referred to [15].

In order to cover the features of all the failure features, we try to use different $L$ and $N$ for the feature extraction. The feature of failure and normal event can not be learned well with a small $L$, the training time will be longer with the smaller $N$. In compromise, for the case 1, we chose the $L=$106 and $N=$20, for the case 2, the $L=$20 and $N=$5 is chosen. It is noted that all these parameters are experienced values with considerations with both feature resolution and processing efficiency, which are a tradeoff but not meant to be optimal.

For a time window data, the learning model from the raw data may lack some information in the time domain of frequency domain. Therefore, the feature extraction module is added, and the feature extraction module focuses on different domains, including original data (time windows without any processing), time domain information extraction based on statistics, frequency domain information extraction based on FFT (Fast Fourier Transformation) and time-frequency domain information extraction based on WPT (Wavelet Packet Transform). Then, we will introduce the different feature extraction methods in detail.

$1)$ Time Domain Feature Extraction: Some statistical features can be extracted from the original data (time window data), including mean, max, min, root mean square and so on. Consider $k-th$ $(k = 1, 2,..., n)$ segment of the time windows, Table I stands for the all statistical function, where $l_k$ stands for the timestamp number of the $k-th$ segment, and $x_{ik}$ stands for the $i-th$ sample in the $k-th$ segment.

\vspace{5pt}
\begin{table}[h]
  \centering
  \begin{tabular}{*{2}{c}}
  \hline
  \hline
    Statistical Measure & Definition  \\ \hline
    Mean& $X_1 = \frac{\sum_{i=1}^{l_k} x_{ik}}{l_k}$  \\
    Root Mean Square &  $X_2 = (\frac{\sum_{i=1}^{l_k} {(x_{ik})}^2}{l_k})^{1/2}$    \\
    Max&  $X_3 = max(x_{ik})$  \\
    Min&  $X_4 = min(x_{ik})$  \\
    Median& $X_5 = \frac{x_{\frac{n}{2}} + x_{\frac{n}{2} + 1}}{2}$  \\
    Range& $X_6 = max(x_{ik}) - min(x_{ik})$  \\
    Crest Factor& $X_7 = \frac{max(|x_{ik}|)}{(\frac{1}{l_k}\sum_{i=1}^{l_k} {(x_{ik})}^2)^{1/2}}$  \\
    Impulse Factor& $X_8 = \frac{max(|x_{ik}|)}{\frac{1}{l_k}\sum_{i=1}^{l_k} {|x_{ik}|}}$   \\
    Margin Factor&  $X_9 = \frac{max(|x_{ik}|)}{(\frac{1}{l_k}\sum_{i=1}^{l_k} {\sqrt{|x_{ik}|}})^{2}}$ \\
    \hline
  \end{tabular}
  \caption{Statistical function of the extracted feature.}\label{table 2}
\end{table}

$2)$ Frequency Domain Feature Extraction: we use FFT (Fast Fourier Transform) on each segment of the window data, and extracts the statistical features in the frequency domain, including the same nine statistical features as time domain.

$3)$ Time-Frequency Domain Feature Extraction: WPT (Wavelet Packet Transform) considers the approximation coefficients for decomposition, and provides a lot of information about data. we use WPT on each segment of the window data, and extracts the statistical features in the time and frequency domain, including the same nine statistical features as time domain.

At the same time, to guarantee efficient learning and immediate decision making, we reduce the dimension of the features by feature reduction. we use three feature reduction methods, includes PCA (principal Component Analysis), LDA (Linear Discriminant Analysis) and LLE (Locally Linear Embedding), to reduce the time complexity and space complexity during learning model, and can also remove the noise samples to some extent.

\subsection{EWMOTE Algorithm}
Synthetic oversampling methods generally are efficient to tackle with the imbalance questions. However, in some conditions, some of current oversampling methods may be inaccurate for imbalanced learning. In terms as EMICIL [7], it generate new samples by learning more feature information about minority class examples based on EMI, but it does not consider the importance of minority class samples, resulting in generating an arbitrary some minority class samples. As shown in Fig.2(a), EMICIL would choose base samples from noisy samples (A, B), the generated samples (square) would also be noisy.  In terms of MWMOTE [6], it consider the importance of minority class examples, but it generates the synthetic samples between chosen base samples and any another samples in same cluster, which divides the minority samples into some clusters by Hierarchical Clustering method. The generated samples would be a wrong samples based on cluster. As shown in Fig.2(b), the middle of a minority cluster may exist some majority class samples, which would generate a noisy or wrong sample (square) when the base samples are C and D. MWMOTE avoid generation of wrong and useless examples by choosing base samples, but not during the cluster. Meanwhile, the cluster-based generated samples method just consider the samples in same cluster, and EMICIL generate samples with the distribution of overall minority samples.



\begin{figure}[htbp]
\vspace{0pt}
\begin{center}
  \includegraphics[width=0.5\textwidth]{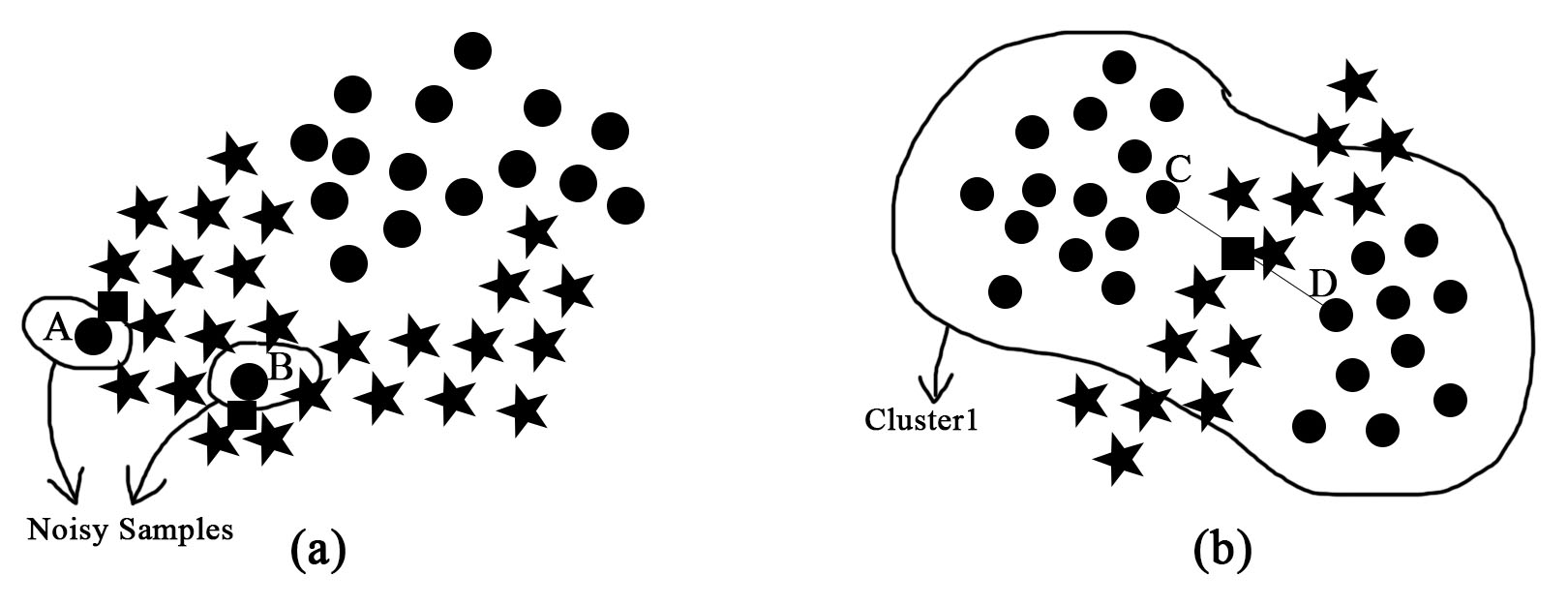}\\
  \caption{(a) Generated samples with noisy samples. (b) Generated samples in a minority class cluster. The stars and circles represent the samples of the majority and minority classes, respectively. Generated samples are shown in square.}\label{Figure 1}
\end{center}
\vspace{-10pt}
\end{figure}

Motivated by the problems stated above, we present a novel minority oversampling technique, EWMOTE, which is based on the imputation of missing values on the minority class examples and weighted minority oversampling technique. In other words, it is based on EMI  (Expectation Maximization Imputation) and MWMOTE. Firstly, We use MWMOTE to identify the most important and hard-to-learn minority class samples $S_{imin}$ from the original minority set. Secondly, we give each number of $S_{imin}$ a selection weight according to its importance. Finally, we generate the synthetic samples from $S_{imin}$ by inducing an attribute missing and filling the missing data based on EMI. The pseudocode for EWMOTE algorithm is shown in Algorithm 2.

The input is as follows: \textbf{$S_{maj}$}: the majority class samples,\textbf{ $S_{min}$}: the minority class samples, \textbf{$N$}: the number of samples we want to generate. We definite one of the minority class as $S_{min}$, and other minority class and majority class as $S_{maj}$ in multi-class learning problems. Meanwhile, there are also some Varied parameters, \textbf{$k_1$}: the number of neighbors for define noisy minority class samples, \textbf{$k_2$}: Number of majority neighbors for generating weighted minority set, \textbf{$k_3$}: Number of minority neighbors for generating weighted minority set. All definition and description of these values are shown in [6], we can chosen proper values based on it. EWMOTE algorithm blend MWMOTE algorithm and EMI algorithm together, from the Algorithm 2, we can see that step1-8 is the part of MWMOTE algorithm and step 10-16 is the part of the EMI algorithm.

\begin{algorithm}
\caption{EWMOTE algorithm}
\label{alg:2}
    \begin{algorithmic}[1]
        \REQUIRE  $S_{maj}$, $S_{min}$, $N$, $k_1$, $k_2$, $k_3$
        \STATE For each minority example $x_i\in$ $S_{min}$, find the nearest neighbor set, $NN(X_i)$. $NN(X_i)$ includes the nearest $k_1$ neighbors, and we calculate based on euclidean distance.
        \STATE Construct the filtered minority set, $S_{minf}$ by deleting some minority class samples which have no minority example in their neighborhood: $S_{minf} = S_{min}- \left\{ x_i\in NN(X_i) \right\} $,  $NN(X_i)$ contains no minority example
        \STATE For each $x_i\in S_{minf}$, compute the nearest majority set, $N_{maj}(x_i)$. $N_{maj}(x_i)$ includes the nearest $k_2$ majority samples, and we calculate based on euclidean distance.
        \STATE Find the borderline majority set, $S_{bmaj}$, the overall can be indicate as $N_{maj}(x_i)s$
        \STATE For each majority example $y_i \in$ $S_{bmaj}$, compute the nearest minority set, $N_{min}(y_i)s$. $N_{min}(y_i)s$ includes the nearest $k_3$ minority examples, and we calculate based on euclidean distance.
        \STATE Find the informative weighted minority set, $S_{imin}$, the overall can be indicate as $N_{min}(y_i)s$
        \STATE For each $y_i\in$ $S_{bmaj}$ and for each $x_i\in$ $S_{imin}$, compute the information weight, $I_w(y_i,x_i)$
        \STATE For each $x_i\in$ $S_{imin}$, compute the selection weight $S_w(x_i) = \sum_{y_i \in S_{bmaj}} I_w(y_i,x_i)$, and covert $S_w(x_i)$ into selection probability $S_p(x_i)$ according to normalization.
        \STATE Initialize the set, $S_{omin} = S_{min}$ ,$i=0$
        \REPEAT
            \STATE \textbf{$i = i+1$}
            \STATE Select a sample $x$ from $S_{imin}$ according to probability distribution $\left\{ S_p(x_i)\right\}$
            \STATE Randomly select an attribute to induce the missing value on $x$, and get $x_{miss}$
            \STATE Call EMI subroutine to estimate missing values[7]  $x_{imp} = EMI(x_{miss})$
            \STATE Add $x_{imp}$ to $S_{omin}$
        \UNTIL $i = N$
        \ENSURE The oversampled minority set, $S_{omin}$.
    \end{algorithmic}
\end{algorithm}

\section{Experiments and Evaluations}

\subsection{Data Preparation and Experimental Settings}
 In this section, we set experiment to evaluate EWMOTE algorithm and other oversampling techniques in two real datasets. Two scenarios with different imbalance ratios and different types of fault are used to evaluate these techniques in industrial diagnostics and prognostics. The imbalance ratio of use case 1 is about 1:16 (the number of majority is 25768 and minority is 1554), and use case 2 is shown in Table II. Use case 1 is a binary imbalance-class problem, and the use case 2 is a multi-class imbalance learning problem, the highest imbalanced ratio of fault in use case 2 is 1:1070.

 Firstly, we set experiment to choose the base classifier. We extract all the fault and normal events from the 90000th to 149999th timestamp of \#21 wind turbine in use case 1 for testing, the rest events of data are used as training set. According to [31], XGBoost shows competitive performance in recall and accuracy than Random Forrest and Gradient Boosting Decision Tree. So in the following experiment, we use XGBoost as our basic classifier.

%

Then, we select the best feature extraction and feature reduction methods. We perform a 10-fold cross validation to verify the generalization performance of different feature extraction and feature reduction methods in training set of two use cases and select the best combination of feature extraction and feature reduction methods as the baseline results (none-sampling methods).

Finally, we compare the performance of 5 methods including none-sampling and 4 oversampling methods in two use cases.  We perform a 10-fold cross validation to verify the generalization performance of our method in training set of two use cases and show the prediction performance of fault in use case 1. We use XGBoost as our base classifier, the number of training epochs is randomly set to 300, and the learning rate to 0.3. For SMOTE, the value of the nearest neighbors, $k$, is set to 5 according to [5]. Meanwhile, for EWMOTE and MWMOTE, the values for different parameters are set as $k_1=5$ (Number of neighbors used for predicting noisy minority class samples), $k_2=3$ (Number of majority neighbors used for constructing informative minority set), $k_3=|S_{min}|/2$ (Number of minority neighbors used for constructing informative minority set), $C_p=3$ (It is used to tune the size and number of clusters), $C_f(th)=5$, and $CMAX = 2$ (The last two are user defined parameters to compute Closeness), according to [6]. It is noted that these parameters are not meant to be optimal but should be kept consistent. According to [8], the number of synthetic samples is set to the same as majority class samples. Referred to the evaluation metrics in [31], we obtain the accuracy, recall, $FAM$ (the average of F-measure, AUC and MCC) and Score for use case 1, and accuracy, recall and $FAM$ for use case 2.

 \newcommand{\tabincell}[2]{\begin{tabular}{@{}#1@{}}#2\end{tabular}}


\vspace{5pt}
\begin{table}[h]
  \centering
  \begin{tabular}{*{8}{c}}
  \hline
  \hline
    Class & N & F1 & F2 & F3 & F4 & F5 & F6  \\ \hline
    Size  &  77043 & 5502 & 4268 & 3820 & 72 & 723 & 21774  \\ \hline
\end{tabular}
\caption{The detail class description of plant failure detection data set, class is the type of plant state. Size is the number of examples.}\label{table 2}
\end{table}


\subsection{Results and Evaluation}

\textbf{Use Case 1: Wind Turbine Freezing Failure Forecast.}

We select the best feature extraction and feature reduction methods, and evaluate the oversampling techniques for imbalanced learning on prediction of fault in use case1.

\begin{figure}[htbp]
\vspace{0pt}
\begin{center}
  \includegraphics[width=0.5\textwidth]{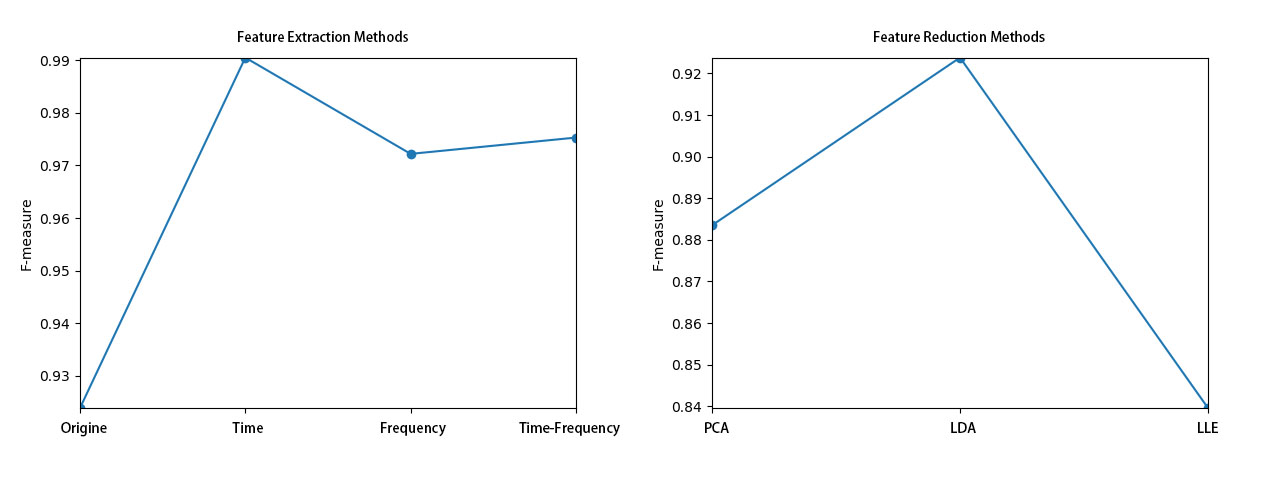}\\
  \caption{The F-measure of different feature extraction and feature reduction methods in use case 1.}\label{Figure 1}
\end{center}
\vspace{-10pt}
\end{figure}

Firstly, We compare the state-of-the-art methods about feature extraction and feature reduction in use case 1. Fig.3 shows the F-measure results of different feature extraction and feature reduction methods. Feature extraction methods include Origin (time window data), Time domain, Frequency domain (FFT) and Time-Frequency domain (WPT). These feature extraction methods are ranked as Time domain, Time-Frequency domain, Frequency domain and Origin in F-measure. Then, Based on the best feature extraction method, Time Domain, the features are dimensioned. When the number of dimensions in the PCA is $n = 75$, it can represent more than $95\%$ of the information, so the dimension is reduced to $n = 75$. Similarly, when LDA and LLE are reduced to 100, more than $95\%$ of the information can be expressed. By feature reduction, the F-measure value is reduced by $11\%$, $6\%$, $15\%$, respectively. Although the feature reduction has a lower F-measure, it is still necessary to choose a proper feature reduction method for the efficiency of model learning and the noise problem of the samples. The fig.3 shows the F-measure of LDA is highest, and the LLE is the lowest. So, we choose Time domain feature extraction method and LDA feature reduction method for the following verification of oversampling methods.

\vspace{5pt}
\begin{table}[h]
  \centering
  \begin{tabular}{*{6}{c}}
  \hline
  \hline
    Method & Recall & Precision & FAM \\ \hline
    None & 0.5385 & 0.9167 &0.7326 \\
    SMOTE & 0.7692 & 0.6918 & 0.7595 \\
    EMICIL & 0.9161 & 0.7939 & 0.8592\\
    MWMOTE & 0.8392 & 0.8633 & 0.8522 \\
    \textbf{EWMOTE} & $\textbf{0.8671}$ & $\textbf{0.8493}$  &\textbf{0.8662}\\
    \hline
  \end{tabular}
  \caption{The Precision, Recall and FAM of compared methods in use case 1.}\label{table 2}
\end{table}

\begin{figure}[htbp]
\vspace{0pt}
\begin{center}
  \includegraphics[width=0.35\textwidth]{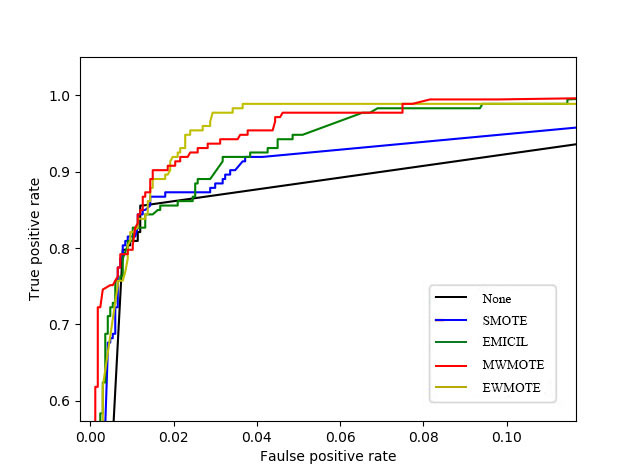}\\
  \caption{The ROC curve comparisons between EWMOTE and other methods in use case1.}\label{Figure 2}
\end{center}
\vspace{-10pt}
\end{figure}

Secondly, we compare the different oversampling methods techniques for imbalanced learning on classification of fault in use case 1. Table III summarized the results of None-sampling, SMOTE, EMICIL, MWMOTE and EWMOTE on the wind turbine dataset, the results include recall, precision and FAM of each method. The ROC graphs are shown in Fig.4. From the results, EWMOTE performs better than None-sampling, SMOTE, EMI and MWMOTE on precision and FAM evaluation metrics. However, the EMICIL methods perform better than EWMOTE on recall. EMICIL perhaps generate some wrong faulty samples so that enlarge the faulty class region and make some generated faulty samples fall inside the normal class region, which can improve the recall owing to identifying more minority class samples, and precision, FAM are reduced owing to the some generated minority class samples fall inside the majority region. At the same time, EWMOTE, which generate new samples based on EMICIL, performs better than SMOTE in all evaluation metrics and MWMOTE in recall and FAM. Generated samples based on EMICIL are more dense and closer to the base samples than SMOTE and MWMOTE.

%


\vspace{5pt}
\begin{table}[h]
  \centering
  \begin{tabular}{*{6}{c}}
  \hline
  \hline
    Method & FN & FP & Score \\ \hline
    None & 943 & 865 & 76.17 \\
    SMOTE & 8673 & 436 & 87.03 \\
    EMICIL & 7628 & 296 & 90.99\\
    MWMOTE & 4036 & 349 & 89.96 \\
    \textbf{EWMOTE} & $\textbf{6826}$ & $\textbf{241}$  &\textbf{92.60}\\
    \hline
  \end{tabular}
  \caption{The FN, FP and Score of compared methods in use case 1.}\label{table 2}
\end{table}

Finally, we compare the oversampling techniques for imbalanced learning on prediction of fault in the testsets. Table IV shows that the FN, FP and Score of five methods on the wind turbine dataset. From the results, EWMOTE performs better than other methods in terms of FP and Score, which select more important base samples and generate new samples based on EMICIL, which avoid generation of wrong samples and generated samples more dense and closer to the base samples than other methods. FN of SMOTE and EMICIL is worse than other methods, which generate new minority class samples wrongly so that enlarge the minority class region and make more majority class examples can not be recognized. However, FN of None-sampling and MWMOTE is better than EWMOTE. In terms of None-sampling, more normal event can be detect owing to the imbalance examples, which make the classifier bias the majority class samples and resulting in a higher prediction error for the fault events. In term of MWMOTE, MWMOTE generate new samples based on cluster and distribution of samples are loose and irregular, EWMOTE generate new samples based on EMI and consider more feature about minority class examples. That's why EWMOTE has a better FP and worse FN than MWMOTE.

\textbf{Use Case 2: Industrial Plant Failure Detection.}

We select the best feature extraction and feature reduction methods, and evaluate the oversampling techniques for imbalanced learning on prediction of fault in use case 2.

\begin{figure}[htbp]
\vspace{0pt}
\begin{center}
  \includegraphics[width=0.5\textwidth]{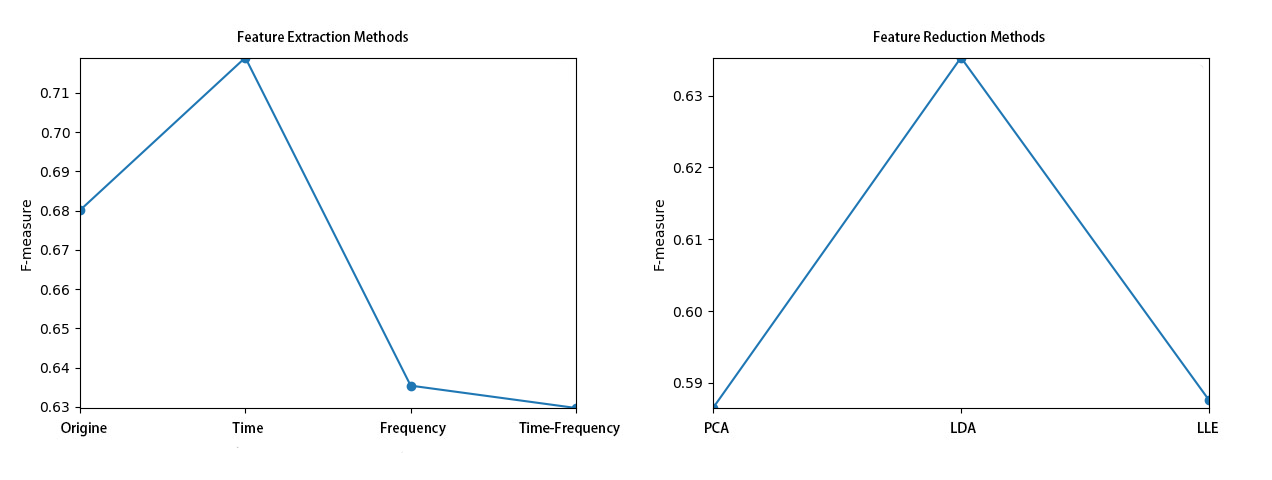}\\
  \caption{ The F-measure of different feature extraction and feature reduction methods in use case 2.}\label{Figure 1}
\end{center}
\vspace{-10pt}
\end{figure}

Firstly, We compare the feature extraction and feature reduction methods in use case 2. Fig.5 shows the F-measure results of different feature extraction and feature reduction, the methods are same as the use case 1. These feature extraction methods are ranked as Time domain, Origin, Frequency domain and Time-Frequency domain in F-measure, and the F-measure of best method is $71.90\%$. Then, based on the best feature extraction method, Time Domain, the feature are dimensioned by PCA, LDA and LLE method, the dimensions of feature reduction is 75, 100, 100,  respectively. We can see that the F-measure is reduced after the feature reduction, owing to the F-measure is not very high after feature extraction, and feature reduction makes it lower, we only perform feature extraction and not reduce the feature dimension.

\vspace{5pt}
\begin{table}[h]
  \centering
  \begin{tabular}{*{5}{c}}
    \hline
    \hline
    Method & Recall & Precision & FAM  \\ \hline
    None & 0.6732 & 0.8205 &0.7146\\
    SMOTE & 0.7934 & 0.8473 & 0.8083  \\
    EMICIL & 0.8468 & 0.8233 & 0.8241  \\
    MWMOTE & 0.8152 & 0.8376 &  0.8157 \\
    \textbf{EWMOTE} & $\textbf{0.8302}$ & $\textbf{0.8573}$  & $\textbf{0.8326}$ \\
    \hline
  \end{tabular}
  \caption{The Precision, Recall and $FAM$ of compared methods in use case 2.}\label{table 2}
\end{table}

Secondly, we compare the different oversampling methods techniques for imbanlanced learning on classification of fault in use case 2. Table V summarized that the results of sampling methods on the industrial plant dataset, the results include average of recall, precision and $FAM$ of each method, and show that EWMOTE method also performs well on imbalanced multi-class learning in terms of precision and F-measure, which illustrate that chose important base samples and generate new minority class samples based on EMI can benefit learning. The results are similar to the results of wind turbine dataset. However, the recall of SMOTE and MWMOTE are worse than EWMOTE, SMOTE and MWMOTE can increase noise samples owing to generating useless minority class samples wrongly to another minority class, which can which can make some other class misclassification and decrease the recall. EMICIL generate new minority class samoles based on EMI, and make a better recall, but worse precision. EWMOTE, which based on the EMICIL and MWMOTE, and performs less worse than EMICIL in recall, but more better in precision. EWMOTE generates samples based on EMI, which consider the distribution of minority samples and less damage to the distribution of the original minority and majority class samples. MWMOTE make the decision boundary closer to majority class region with the samples generating, which results worse recall and better precision.

\vspace{5pt}
\begin{table}[h]
  \centering
  \begin{tabular}{*{5}{c}}
    \hline
    \hline
    Class & Recall & Precision & FAM \\ \hline
    F1 & 0.7836  & 0.8273  &  0.8002\\
    F2 & 0.8855 & 0.8203 & 0.8490\\
    F3 & 0.9031 &  0.8603 &  0.8793\\
    F4 & 0.8571 & 1.0 & 0.9244 \\
    F5 & 0.5417 & 0.6290 & 0.5817\\
    F6 & 0.8787 & 0.9105& 0.8822\\
    Normal & 0.9614 & 0.9536 & 0.9116 \\
    \hline
  \end{tabular}
  \caption{The Precision, Recall, F-measure and MCC of different classes based on EWMOTE in use case 2}\label{table 2}
\end{table}

Meanwhile, Table VI shows the classification result of each class in EWMOTE method, including normal and other 6 fault classes, shows that our proposed generated minority class samples technique performs best in overall result. In terms of low imbalance ratio class (F6) and medium imbalance ratio class (F1-F3), we can see that EWMOTE perform better, but a little worse in a high imbalance ratio class (F4 and F5), F4 performs better, but F5 worse. Figure 6 illustrates the EWMOTE performs more robust than other synthetic sampling methods on identify multi-class faulty samples with different imbalanced ratios, especially on faulty types with high imbalanced ratios. We can find that for F4 the performances are close between EWMOTE, MWMOTE and EMICIL. However, for F5, the True Positive rate of EWMOTE is higher than MWMOTE, while the False Positive rate is lower than EMICIL. These phenomenon illustrates that some improvement can be done for EWMOTE algorithm in the condition of high imbalance ratio.

\begin{figure*}
\begin{minipage}{0.18\linewidth}
  \centerline{\includegraphics[width=1.8in]{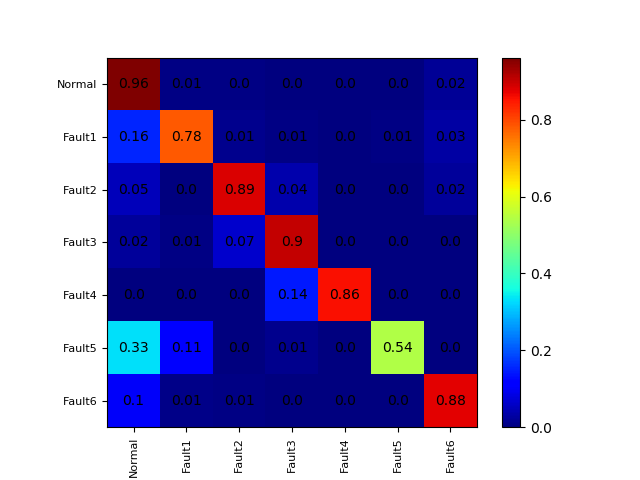}}
  \subcaption{EWMOTE}

\end{minipage}
\hfill
\begin{minipage}{0.18\linewidth}
  \centerline{\includegraphics[width=1.8in]{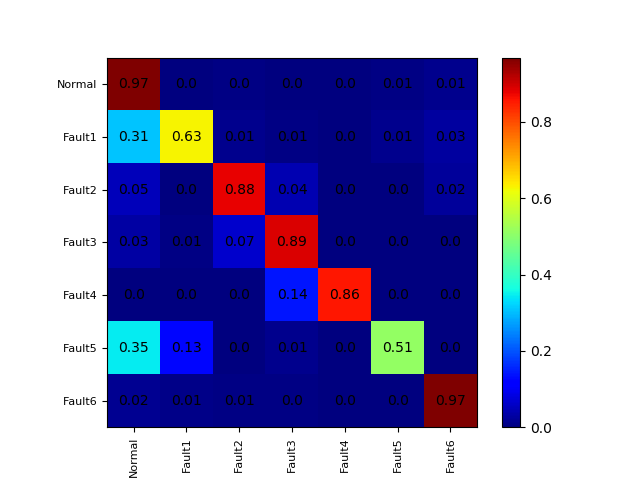}}
  \subcaption{MWMOTE}
\end{minipage}
\hfill
\begin{minipage}{0.18\linewidth}
  \centerline{\includegraphics[width=1.8in]{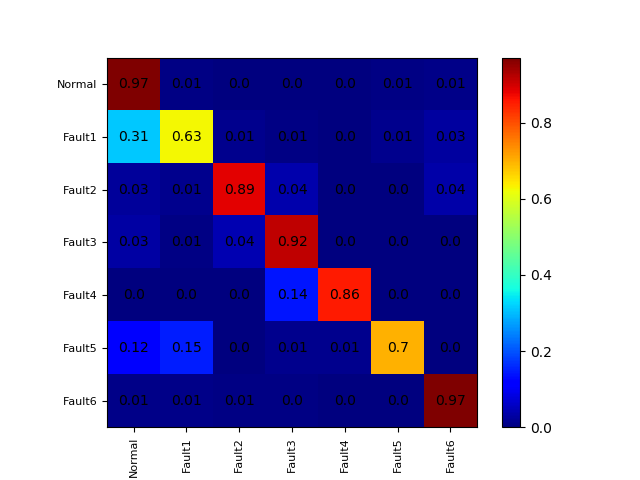}}
  \subcaption{EMICIL}
\end{minipage}
\hfill
\begin{minipage}{0.18\linewidth}
  \centerline{\includegraphics[width=1.8in]{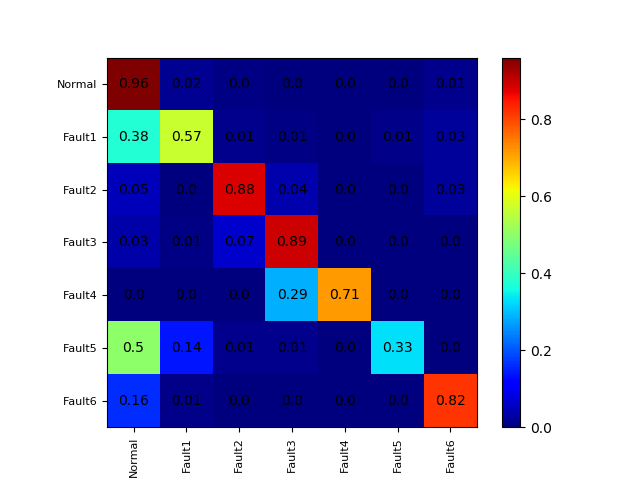}}
  \subcaption{SMOTE}
\end{minipage}
\hfill
\begin{minipage}{0.18\linewidth}
  \centerline{\includegraphics[width=1.8in]{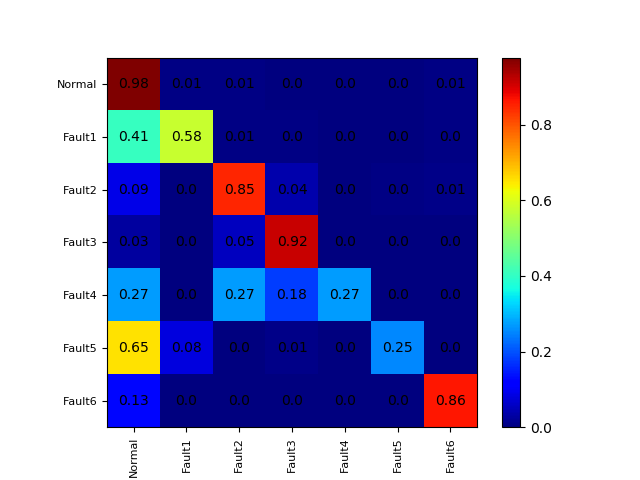}}
  \subcaption{None Sampling}
\end{minipage}
\caption{Confusion matrix of EWMOTE (\textbf{a}); MWMOTE (\textbf{b}); EMICIL (\textbf{c}); SMOTE (\textbf{d}) and None Sampling (\textbf{e}).}
\label{Figure 6}
\end{figure*}

\section{Conclusion and Future Work}


In this paper, we proposed a general scheme of tackling the imbalanced fault diagnostics and prognostics, which includes the Feature Extraction, Feature Reduction, Data-level methods unit and Learning unit. Meanwhile, we proposed a novel synthetic oversampling technique (EWMOTE) aims to overcome the challenge about generating wrong and unnecessary samples, our technique decrease imbalance ratio by generating new minority class samples based on weighted minority and imputation of missing samples. Furthermore, this technique predict the freezing failure of wind turbine and normal and different faults of industrial plant based on two popular PHM competitions. The experiments show that EWMOTE performs better than other None-sampling and oversampling methods on not only binary but also multi-faults classifications. To improve the efficiency and effectiveness of the prognostic model for industrial Systems, a good data segmentation and some other feature extraction methods could be further studied. Meanwhile, a new ensemble method based on some algorithm-level methods or some undersampling methods could be studied in the future. Finally, EWMOTE involves some mutative parameters, which can be optimized and result can be improved according to the specific scenes and problem at hand.


%

\appendices

\ifCLASSOPTIONcaptionsoff
  \newpage
\fi

\end{document}